\documentclass[conference]{IEEEtran}
\IEEEoverridecommandlockouts
\usepackage{cite}

\usepackage{amsmath,amssymb,amsfonts}
\usepackage{algpseudocode}
\usepackage{graphicx}
\usepackage{booktabs}
\usepackage{multirow}
\usepackage{textcomp}
\usepackage{bm}
\usepackage{xcolor}
\usepackage[colorlinks=trn ue, linkcolor=blue, citecolor=red, urlcolor=cyan]{hyperref}
\def\BibTeX{{\rm B\kern-.05em{\sc i\kern-.025em b}\kern-.08em
    T\kern-.1667em\lower.7ex\hbox{E}\kern-.125emX}}
\begin{document}
\definecolor{cvprblue}{rgb}{0.21,0.49,0.74}
\title{EvFlow-GS: Event Enhanced Motion Deblurring with Optical Flow for 3D Gaussian Splatting

\thanks{
\dag: Corresponding author.}
}
\DeclareRobustCommand*{\IEEEauthorrefmark}[1]{%
    \raisebox{0pt}[0pt][0pt]{\textsuperscript{\footnotesize\ensuremath{#1}}}}

	\author{
    
		\IEEEauthorblockN{
			Feiyu An\IEEEauthorrefmark{1}, 
			Yufei Deng\IEEEauthorrefmark{1},
			Zihui Zhang\IEEEauthorrefmark{1},
			Rong Xiao\IEEEauthorrefmark{1\dag}}
\IEEEauthorblockA{
    \IEEEauthorrefmark{1}College of Computer Science, Sichuan University, Chengdu, China
    \\
    \{anfeiyu, yufeideng, zhangzihui\}@stu.scu.edu.cn, rxiao@scu.edu.cn
}
	}
    
\maketitle

\begin{abstract}
Achieving sharp 3D reconstruction from motion-blurred images alone becomes challenging, motivating recent methods to incorporate event cameras, benefiting from microsecond temporal resolution.
However, they suffer from residual artifacts and blurry texture details due to misleading supervision from inaccurate event double integral priors and noisy, blurry events.
In this study, we propose EvFlow-GS, a unified framework that leverages event streams and optical flow to optimize an end-to-end learnable double integral (LDI), camera poses, and 3D Gaussian Splatting (3DGS) jointly on-the-fly. 
Specifically, we first extract edge information from the events using optical flow and then formulate a novel event-based loss applied separately to different modules.
Additionally, we exploit a novel event-residual prior to strengthen the supervision of intensity changes between images rendered from 3DGS.
Finally, we integrate the outputs of both 3DGS and LDI into a joint loss, enabling their optimization to mutually facilitate each other. 
Experiments demonstrate the leading performance of our EvFlow-GS.

\end{abstract}

\section{Introduction}
\label{1}
Neural Radiance Fields (NeRF)~\cite{mildenhall2020nerf} and 3D Gaussian Splatting (3D-GS)~\cite{kerbl3Dgaussians} employ implicit neural representations and explicit Gaussian primitives, respectively, for 3D reconstruction, yet both rely on blur-free, multi-view images captured under well-controlled conditions to achieve high-quality reconstruction. However, such ideal capture conditions are often unattainable in real scenes. In low-light settings, cameras require extended exposure to gather sufficient illumination, during which any slight movement inevitably results in motion blur.
Recent works \cite{zhao2024badgaussians, lee2025comogaussian,Lee_2023_CVPR,lee2024deblurring} have shown that 3D representations can be recovered from blurry multi-view images. However, reconstruction relying solely on blurry images remains ill-posed and yields suboptimal visual quality, because such inputs lack essential motion information.

\begin{figure}[t]
    \centering
    \includegraphics[trim=47 320 820 155, clip, width=0.48\textwidth]{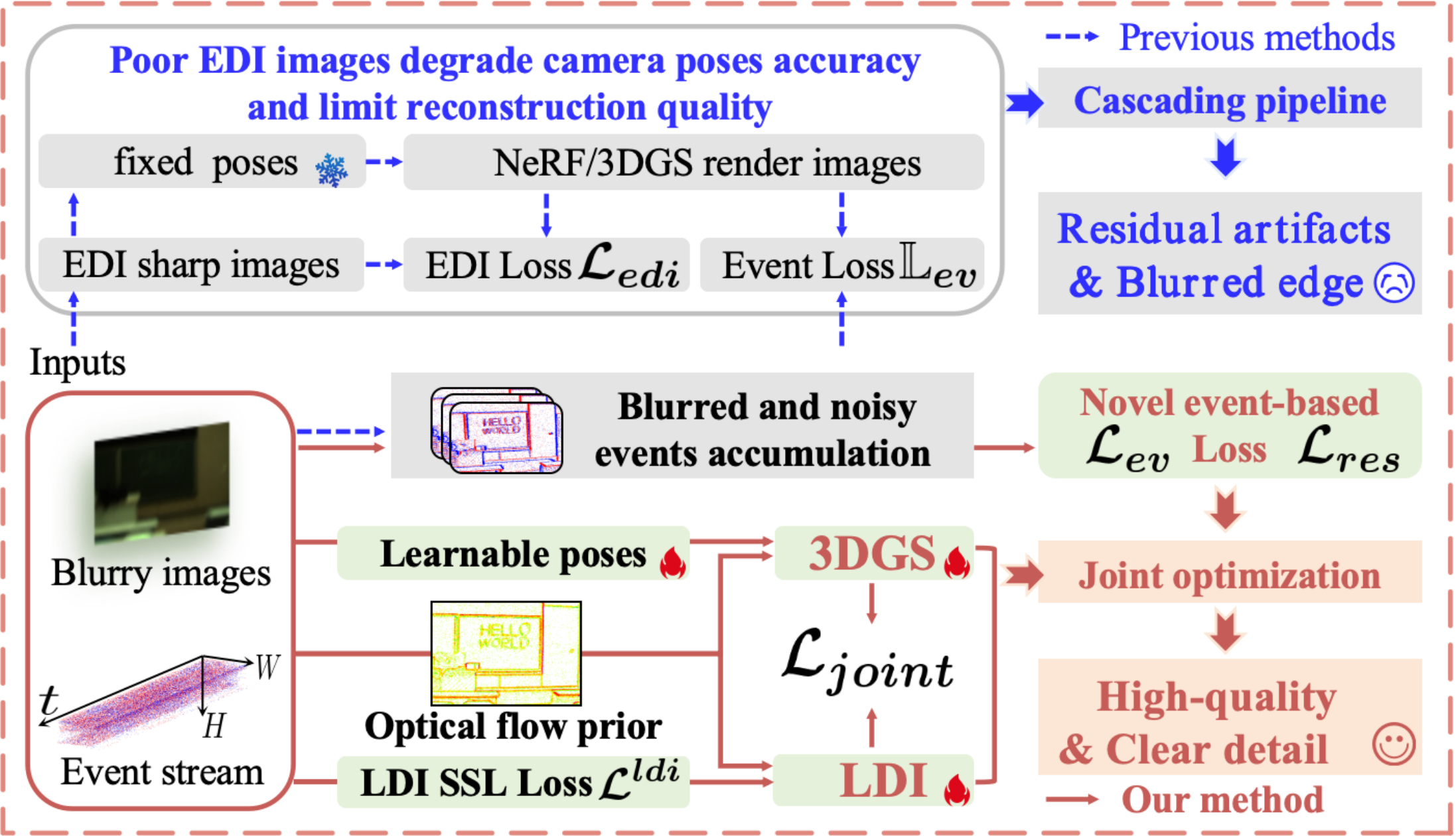}
    \caption{Comparison between our method and the previous methods. Our method integrates LDI, pose estimation, and 3DGS into a unified framework that enables their co-optimaization signified by (\includegraphics[height=0.8em]{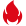}).}
    \label{fig:1}
\vspace{-\baselineskip}
\end{figure}

Benefiting from the high temporal resolution of event cameras, several works have explored combining motion-blurred images with events to facilitate 3D reconstruction. E2NeRF\cite{qi2023e2nerf} and following works~\cite{lee2025diet,qi2024deblurring,Cannici_2024_CVPR, deng2025ebadgaussianeventdrivenbundleadjusted} model camera motion during exposure through blurry image formation~\cite{zhao2024badgaussians} and exploit continuous event-by-event supervision \(\mathbb{L}_{ev}\) to improve reconstruction results, as shown in Fig.~\ref{fig:1}. EvDeblurNeRF\cite{Cannici_2024_CVPR} and DiET-GS\cite{lee2025diet} further leverage the Event Double Integral (EDI)~\cite{Pan2018BringingAB} as a color deblurring prior to provide additional color constraint \(\mathcal{L}_{edi}\).

\label{sec:intro}
Nevertheless, these reconstruction approaches still suffer from residual artifacts and blurry texture details, primarily due to the following three factors: \textbf{(i) Inaccurate EDI Priors}. The physical estimation of EDI is inaccurate due to the unknown contrast threshold and large event noise. The discrepancy in the double integration calculation leads to considerable noise in EDI~\cite{qi2024e3nerfefficienteventenhancedneural}. Existing studies utilizing EDI can be broadly categorized into two types. First, E2NeRF~\cite{qi2023e2nerf} and DiET-GS~\cite{lee2025diet} first employ the EDI physical model to generate deblurred images, then estimate fixed camera poses, and finally construct a 3D representation, as shown in Fig.~\ref{fig:1}. However, due to the low quality of the initial EDI images, such cascaded pipelines introduce significant error accumulation, which damages the accuracy of camera pose estimation and limits the fidelity of scene reconstruction. Second, EvDeblurNeRF~\cite{ Cannici_2024_CVPR} which only uses them to supervise the model leads to obvious unwanted color artifacts and loss of high-frequency details in the 3D rendering, particularly when EDI retains substantial noise. \textbf{(ii) Blurred Events.} Existing methods~\cite{Cannici_2024_CVPR,lee2025diet,qi2023e2nerf,qi2024deblurring,deng2025ebadgaussianeventdrivenbundleadjusted} first convert the continuous event stream into discrete event frames to construct event loss \(\mathbb{L}_{ev}\). However, creating these frames requires accumulating events, which introduces high latency. In addition, events are considered blurred when those originating from the same moving edge are misaligned in space and time, appearing at different pixel locations and timestamps~\cite{paredes2020back}. Using accumulated events as supervision signals for rendering images introduces erroneous and blurry edge trajectories, leading to conspicuous edge ghosting artifacts in rendering. \textbf{(iii) Log-Intensity-based Supervision}. Events provide differential log-brightness changes, while 3DGS renders absolute brightness. Previous works~\cite{Cannici_2024_CVPR,lee2025diet,qi2023e2nerf,qi2024deblurring,deng2025ebadgaussianeventdrivenbundleadjusted} supervise the model by comparing the predicted log-intensity changes with event streams. Although consistent with the Event Generation Model (EGM), this way suffers from four critical issues. First, event streams contain significant intrinsic noise, especially in low-intensity regions. Establishing log-intensity-based supervision \(\mathbb{L}_{ev}\) can amplify noise and introduce spurious intensity changes in non-edge regions. Second, the threshold in the event generation model varies across polarity, space, and time~\cite{Pan2018BringingAB}. Nevertheless, previous works assume a constant threshold, resulting in degraded supervision quality. Third, log-intensity-based supervision \(\mathbb{L}_{ev}\) magnifies relative intensity changes, so small deviations can produce incorrect predictions of intensity changes and degrade rendering quality in low-intensity regions. Fourth, using these differential operations relies on a uniform-motion assumption that is ill-suited to real-world scenarios\cite{deng2025ebadgaussianeventdrivenbundleadjusted}.

\begin{figure*}[!htbp]
    \centering
    \includegraphics[trim=85 132 15 207, clip, width=\textwidth]{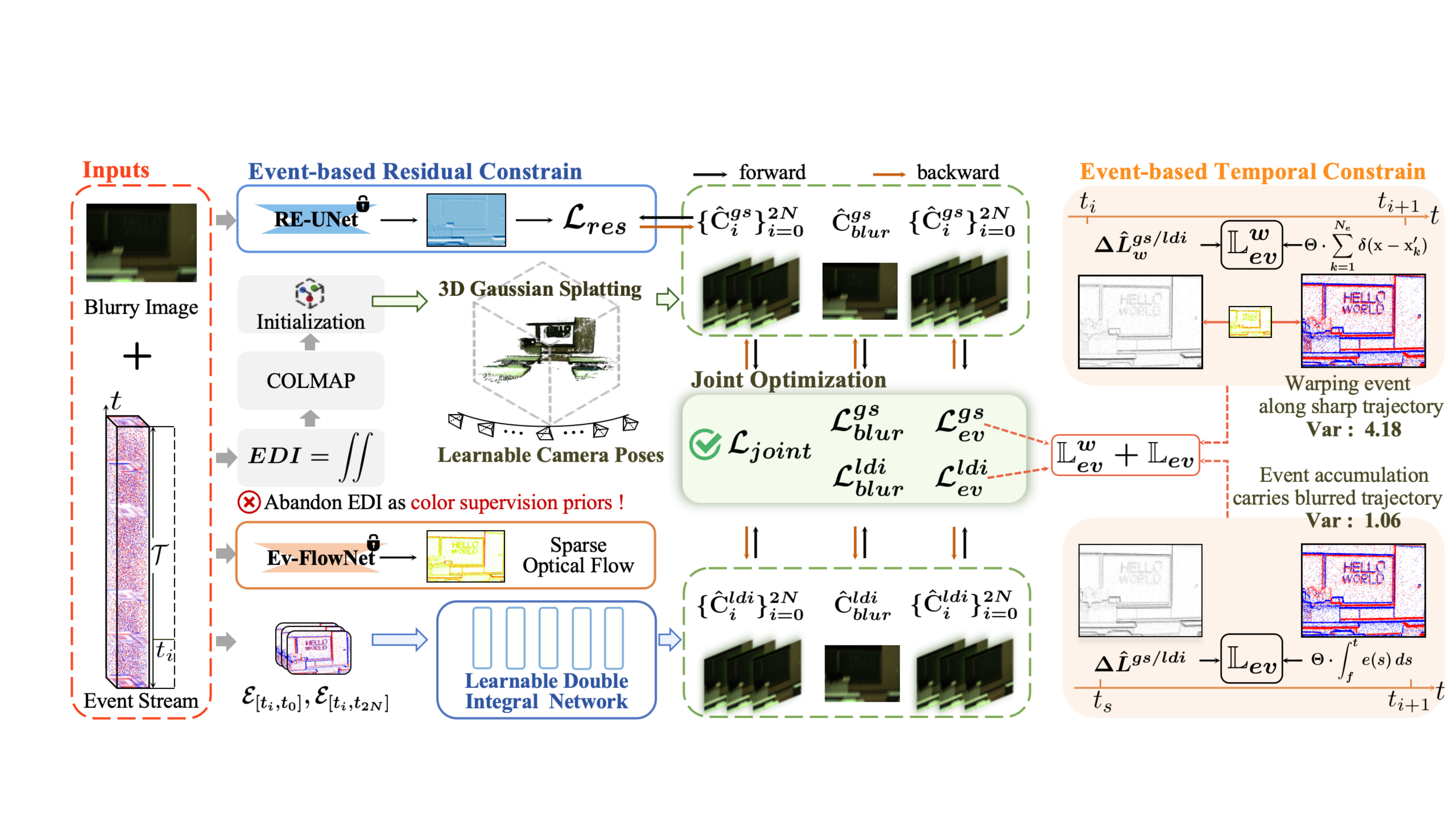}
    \caption{\textbf{Overall framework of our EvFlow-GS.} We propose three novel strategies to effectively leverage event stream for optimizing deblurring 3DGS.}
    \label{fig:2}
\vspace{-\baselineskip}
\end{figure*}

To address the previously discussed problems, we propose \textit{\textbf{EvFlow-GS}}, a unified framework that leverages event streams and optical flow to optimize an end-to-end learnable double integral (LDI), camera poses, and 3DGS jointly on-the-fly, illustrated in Fig.~\ref{fig:2}. \textbf{First, to avoid constraining our model within a cascading framework and to mitigate the impact of poor EDI,} we propose a cooperative framework that jointly integrates LDI network, learnable pose estimation, and 3DGS within a unified optimization architecture. The LDI network achieves self-supervision through $\mathcal{L}_{ldi}$, which is similar to the supervision in 3DGS. Finally, we incorporate a joint loss $\mathcal{L}_{\text{joint}}$ that integrates outputs from both 3DGS and LDI to enable mutual enhancement and unified optimization of each other.
\textbf{Second, to reduce the impact of blurred events and better match non-uniform motion}, we warp events along candidate spatio-temporal trajectories to align them, an operation that provides a supervision signal whose sharpness is maximized. We then leverage this sharp edge signal to design $\mathbb{L}_{ev}^w$, which supervises different modules that better match the correct edge trajectory.
\textbf{Third, to mitigate the impact caused by log-intensity-based supervision}, we redesign the event-based loss $\mathcal{L}_{res}$ using a novel residual directly based on inter-frame intensity differences. This residual, extracted from a pretrained model and used as a prior, provides a cleaner supervision signal that combines rich motion information from events with fine texture cues from blurry images, enabling the model to recover finer details with fewer artifacts. We evaluate our method on one synthetic and two real-world datasets. Experimental results demonstrate that our method outperforms existing baselines. In summary, our contributions are as follows: Experimental results demonstrate that our method outperforms existing baselines. In summary, our contributions are as follows: 
\begin{itemize}
    \item We present a unified framework that jointly optimizes LDI, pose estimation, and 3DGS. Through joint optimization of the branches, we improve both texture quality and color accuracy of 3D reconstruction.
    \item We utilize optical flow and a novel event-driven residual prior to propose novel loss functions that supervise the different modules of the model, enhancing robustness against blurred and noisy events.
    \item Experimental results demonstrate that our method outperforms previous baselines, exhibiting fewer artifacts.
\end{itemize}

\section{Preliminaries}
\noindent\textbf{Event Generation Model.}\hspace{1em}Event cameras trigger events when the logarithmic brightness change between \(t_k\) and \(t_{prev}\) exceeds a threshold \(\Theta\). Each event tuple \( e_k = (t_k, x_k, y_k, p_k)\) satisfies 
\(p\Theta = \log I(t_k, \bm{x}) - \log I(t_{\text{prev}}, \bm{x})\), where \(\Theta > 0\), $I(t_k, \bm{x})$ is brightness at pixel $\bm{x}_k = (x_k, y_k)^T$ and $p \in \{+1, -1\}$ encodes change direction. Over time window $\Delta t = [f, t]$, events give brightness integral:
\begin{equation}
\Theta \cdot \int_{f}^{t} e(s) \, ds = \log I(t, \bm{x}) - \log I(f, \bm{x})
\label{eq:1}
\end{equation}
where $e(t_k) = p \cdot \delta(t_k - t_{\text{prev}})$ is the continuous expression of the event, $\delta(\cdot)$ is Dirac delta. Defining $L(t, \bm{x}) = \log(I(t, \bm{x}))$, we obtain $\Delta L(f, t) = \Theta \cdot \int_{f}^{t} e(s) \, ds$. Under the assumptions of Lambertian surfaces, constant illumination, and small $\Delta t$, Eq. \eqref{eq:1} can be simplified to derive the event-based photometric constancy constraint:
\(
  \Delta L(f, t) \approx - \nabla L(f, \bm{x}) \cdot \bm{u}(\bm{x}) \Delta t
\label{eq:2}
\)
which states that events are triggered by the spatial gradients of the brightness signal, \( \nabla L = (\delta_x L, \delta_y L)^T \), moving with optical flow $\bm{u}$. Event rate reaches maximum when $\bm{u} \parallel \nabla L$ and drops to zero when $\bm{u} \perp \nabla L$.\\
\textbf{Event-based Motion Deblurring Model.}\hspace{1em}A motion-blurred image is generated by averaging the latent sharp images over the exposure time: $B = \frac{1}{T} \int_{t \in \mathcal{T}} I(t, \bm{x}) dt$ with \( T \) denoting the duration of exposure period \( \mathcal{T} \). Combining Eq. \eqref{eq:1} with the blur model, we can obtain the relation between blurry frame \( B \) and latent image \( I(f) \):
\begin{equation}
I(f)
=
\frac{B}{E(f, \mathcal{T})}
\label{eq:3}
\end{equation}
where $f \in \mathcal{T}$ and $E(\cdot)$ denotes the event double integral, defined as
\(E(f, \mathcal{T})
=
\frac{1}{T}
\int_{t \in \mathcal{T}}
\exp\!\left(
\Theta \int_{f}^{t} e(s)\, ds
\right) dt \).
Equation \eqref{eq:3} is also known as EDI\cite{Pan2018BringingAB}. Previous research \cite{Cannici_2024_CVPR, lee2025diet, qi2024e3nerfefficienteventenhancedneural} has commonly leveraged Eq. \eqref{eq:3} to recover a sequence of sharp images, which are used to formulate an EDI prior loss against the rendered images from NeRF or 3DGS. In the following, instead of relying on the EDI loss, we propose a unified framework that jointly optimizes a lightweight, trainable neural network and 3DGS.
\section{Methods}
\subsection{Trajectory Modeling in 3DGS}\label{sec:4.1}
We aim to reconstruct the scene using an ensemble of Gaussian primitives. Inspired by \cite{zhao2024badgaussians}, we employ Bézier interpolation to optimize the continuous trajectory of the camera pose over the exposure interval \( \mathcal{T} \). Specifically, we utilize nine control points \( \mathbf{T}_0, \mathbf{T}_1, \ldots, \mathbf{T}_8 \in \mathbb{SE}(3) \) to construct a Bézier curve. The continuous camera pose \( \mathbf{P}(t) \) at any time \( t \) is expressed as:
\begin{equation}
\mathbf{P}(t_k) = \prod_{j=0}^{8} \exp \left( \binom{8}{j} (1 - \tau)^{8 - j} \tau^j \cdot \log(\mathbf{T}_j) \right)
\label{eq:camera pose}
\end{equation}
where \( \tau = t / T \in [0, 1] \). During training, the camera poses are jointly optimized with the following supervisions. 3D Gaussians are projected onto the image plane under learnable camera poses. The final pixel colors $\{\hat{\mathbf{C}}^{gs}(t_i)\}_{i=0}^{2N}$ are then computed via $\alpha$-blending. 
\vspace{-1pt}
\subsection{Event-based Temporal Constraint}
\label{sec:4.2}

Event cameras provide microsecond, blur-free signals that help 3DGS reconstruct scenes more clearly through the relation in Eq. \eqref{eq:1}.
To fully utilize the high temporal resolution of events, we employ a multi-temporal random sampling strategy.

Specifically, we uniformly sample timestamps $\{t_i\}_{i=0}^{2N}$ over $\mathcal{T}$. Since the right side of Eq. \eqref{eq:1} provides a supervision signal, we define the event-based loss as follows:
\begin{equation}
\mathbb{L}_{ev}^{gs}= \frac{1}{2N} \sum_{i=0}^{2N-1}
\left\|
\frac{\Delta L(t_s,t_{i+1})}{\bigl\|\Delta L(t_s,t_{i+1})\bigr\|_2}
-
\frac{\Delta\hat{L}^{gs}(t_s,t_{i+1})}{\bigl\|\Delta\hat{L}^{gs}(t_s,t_{i+1})\bigr\|_2}
\right\|^{2}
\label{eq:5}
\end{equation}
where $\Delta\hat{L}^{gs}(t_s,t_{i+1})=\log (\hat{\mathbf{C}}^{gs}(t_{s})) - \log (\hat{\mathbf{C}}^{gs}(t_{i+1}))= 
\log \frac{\hat{\mathbf{C}}^{gs}(t_s)}{\hat{\mathbf{C}}^{gs}(t_{i+1})}$ and \( t_s \sim \mathcal{U}(\{t_0, \ldots, t_i\}) \). We normalize the accumulated events to eliminate the effect of unknown $\Theta$. Brightness changes in the event stream offer valuable supervision for recovering sharp details from blurred inputs. However, using blurred event accumulation to supervise 3DGS leads to ghosting artifacts in the edges. 
Benefiting from optical flow that provides a precise estimation of pixel-wise motion, we leverage the estimated optical flow to warp and integrate events over time to recover a sharp representation of brightness changes from events. 

Specifically, we use the optical flow $\bm{u}_i$ estimated by EV-FlowNet \cite{hagenaarsparedesvalles2021ssl} to warp the event stream through
\(
\bm{x}'_k
=
\bm{x}_k
+
\left(t_{i+1} - t_k\right)\,\bm{u}_i(\bm{x}_k)
\)
Following \cite{paredes2020back}, to enable effective supervision of varying motion speeds during exposure, we compute $\Delta\hat{L}$ by warping spatial gradients from $\hat{\mathbf{C}}(t_i)$ to the current time $t_{i+1}$: \(
\Delta \hat{L}^{gs}_w(t_i, t_{i+1}) \doteq -\mathcal{W}_{i}^{i+1}\!\left(\nabla \hat{\mathbf{C}}^{gs}(t_i,\bm{x})\right) \cdot \bm{u}_i(\bm{x})\), where $\mathcal{W}_{i}^{i+1}$ denotes the warping function. The optimized events loss can then be formulated as: 
\begin{equation}
\mathbb{L}_{ev}^{w}
=
\mathbb{L}_{ev}^{gs}
\bigl(
\Delta L_w,\,
\Delta \hat{L}^{gs}_w
\bigr)
\end{equation}
where $ \Delta L_w(t_i, t_{i+1}) \doteq \Theta \cdot\sum_{k=1}^{N_e} \delta(\bm{x} - \bm{x}'_k)$. 

Inspired by~\cite{paredes2020back}, we can use the variance of events $\mathrm{var}(\cdot)$ to assess the accuracy of motion edge trajectory estimation provided by events. Fig.~\ref{fig:2} shows that our warped events become concentrated on a smaller set of pixels, resulting in a large variance. In contrast, the accumulated events without motion compensation exhibit a dispersed distribution, leading to a small variance. This is because our projection of edges onto the reference time plane is performed along a sharp trajectory.
Compared to event loss used in recent methods \cite{lee2025diet, qi2024deblurring, Cannici_2024_CVPR}, our warped-event loss is also not affected by any motion speed. Therefore, our method not only provides higher accuracy for non-uniform motion, and it performs better in edge areas with weaker supervised textures. Finally, the temporal loss function of events is given as: \begin{equation}
\mathcal{L}^{gs}_{ev}
=
\mathbb{L}_{ev}^{w}
+
\mathbb{L}_{ev}^{gs}.
\end{equation}

\begin{table*}[t]
\centering
\caption{Quantitative comparisons on both synthetic and real-world datasets. The results are the average of every scenes within the dataset. The best results are in \textbf{bold} while the second best results are \underline{underscored}.}
\renewcommand{\arraystretch}{1.12}
\setlength{\tabcolsep}{3pt}
\vspace{-2.3mm}
\resizebox{\textwidth}{!}{
\begin{tabular}{c|c|ccc|cccc|cccccc}
\hline
\multirow{1}{*}{Dataset} 
& Metric 
& 3DGS+MPRnet 
& 3DGS+EFnet 
& 3DGS+EDI 
& Deblurring 3DGS  
& CoMoGaussian 
& DP-NeRF
& Bad-Gaussian 
& EvDeblurNeRF
& E2NeRF
& DiET-GS 
& EBAD-NeRF 
& E3NeRF
& EvFlow-GS \\
\hline

\multirow{3}{*}{\begin{tabular}{c} EBAD-NeRF:\\  Real-World\end{tabular}} 
& PSNR↑ & 23.65 & 24.69 & 28.01 & 28.13 & 28.26 & 29.29 & 28.77 & 29.26 & 28.18 & \underline{29.90} & 29.37 & -- & \textbf{32.92}\\
& SSIM↑ & 0.741 & 0.728 & 0.813 & 0.814 & 0.836 & 0.849 & 0.829 & 0.845 & 0.851 & \underline{0.874} & 0.857 & -- & \textbf{0.910} \\
& LILPS↓ & 0.409 & 0.347 & 0.205 & 0.218 & 0.167 & 0.168 & 0.175 & 0.192 & 0.238 & \underline{0.149} & 0.165 & -- & \textbf{0.131} \\
\hline

\multirow{3}{*}{\begin{tabular}{c} Real-World \\ Challenge\end{tabular}}
& PSNR↑ & 27.12 & 28.06 & 28.54 & 30.74 & 30.78 & 28.66 & 26.53 & 28.47 & 29.15 & 31.22 & 28.01 & \underline{31.37} & \textbf{33.14} \\
& SSIM↑ & 0.846 & 0.894 & 0.920 & 0.834 & 0.828 & 0.915 & 0.886 & 0.912 & 0.886 & 0.875 & 0.869 & \textbf{0.942} & \underline{0.928} \\
& LILPS↓ & 0.395 & 0.356 & 0.273 & 0.231 & 0.208 & 0.293 & 0.285 & 0.271 & 0.233 & \underline{0.174} & 0.293 & 0.194 & \textbf{0.139} \\
\hline
\multirow{3}{*}{\begin{tabular}{c} EBAD-NeRF \\ Synthetic\end{tabular}}
& PSNR↑ & 23.11 & 24.35 & 27.98 & 24.59 & 26.20 & 24.32 & 24.56 & 27.96 & 27.10 & 27.43 & \underline{28.53} & -- & \textbf{31.06} \\
& SSIM↑ & 0.665 & 0.778 & 0.813 & 0.671 & 0.773 & 0.718 & 0.739 & 0.864 & 0.812 & 0.846 & \textbf{0.920} & -- & \underline{0.913} \\
& LILPS↓ & 0.387 & 0.290 & 0.204 & 0.285 & 0.261 & 0.310 & 0.292 & 0.186 & 0.251 & 0.192 & \underline{0.166} & -- & \textbf{0.099} \\
\hline

\end{tabular}
}
\label{tab:2}
\vspace{-\baselineskip}
\end{table*}
\vspace{-\baselineskip}
\subsection{Blur Reconstruction Constraint}
To account for real-world non-uniform motion, we reuse the optical flow from Sec.~\ref{sec:4.2} to upsample frames by \(M=4\), inserting three frames between sharp images: \(\tilde{\mathbf{C}}_{i,m}^{gs}(\bm{x})=\tilde{\mathbf{C}}_{i}^{gs}\!\left(\bm{x}+\tfrac{m}{M}\bm{u}_{i}(\bm{x})\right), m=1,\dots,M{-}1\). The blur-consistency loss is defined as: 
\begin{equation}
\mathcal{L}_{blur}^{gs}
=
\mathcal{L}\!\left(
\mathbf{C}_{gt},
\;
\frac{1}{2N M + 1}
\sum_{i=0}^{2N}
\sum_{m=0}^{M-1}
\tilde{\mathbf{C}}_{i,m}^{gs}
\right)
\end{equation}
where \(\tilde{\mathbf{C}}_{i,0}^{gs}\triangleq\tilde{\mathbf{C}}_{i}^{gs}\) and $\mathcal{L}$ consists of an $L_1$ term together with a D-SSIM component, which is defined as $\mathcal{L} = (1 - \lambda_{SSIM})\mathcal{L}_1 + \lambda_{SSIM} \mathcal{L}_{D\text{-SSIM}}$.
\subsection{Event-based Residual Constraint}
\label{sec:4.3}
Although event streams offer blur-free details, their inherent noise often introduces unnatural artifacts in 3DGS. In addition, when $\hat{\mathbf{C}}^{gs}(t_{s})$ is small, even a minor intensity difference between $\hat{\mathbf{C}}^{gs}(t_{s})$ and $\hat{\mathbf{C}}^{gs}(t_{i+1})$ can cause $\Delta\hat{L}^{gs}(t_s,t_{i+1})$ to vary drastically. This leads to unstable training as the loss becomes overly sensitive to slight variations in rendering intensity within low-intensity regions. Inspired by\cite{chen2024motion}, we propose here to exploit a novel residual form $R(k)$, which differs from log-intensity difference from Eq. (\ref{eq:2}), to further constrain 3DGS training in low-intensity regions.

In particular, we redefine the residual as $R(t_i) = I(t_i) - I(t_N)$ and first compute the aligned event accumulation $\mathcal{E}_{[t_N,t_i]}=\int_{t_N}^{t_i}e(s)\,ds$. We use the pretrained RE-UNet from~\cite{chen2024motion} to calculate the residual: $R(t_i)=\mathcal{G}(\mathcal{E}_{[t_N,t_i]}, \mathbf{C}_{gt})$ for $i\in\{0,\ldots,2N\}$. Unlike noise event accumulation, the pre-trained model utilizes the blurry image $\mathbf{C}_{gt}$ as a structural prior to filter out noise inconsistent with the scene while preserving genuine intensity changes consistent with image content. The resulting residual exhibits a higher signal-to-noise ratio. Using this novel residual to supervise 3DGS avoids the sensitivity of log-intensity residual in dark regions and leads to more stable training. We construct an event-based residual loss to improve 3DGS's robustness against event noise: 
\begin{equation}
\mathcal{L}_{\text{res}} = \frac{1}{2N+1}\sum_{i=0}^{2N} \| R(t_i) - \hat{R}(t_i) \|^2
\end{equation}
where $\hat{R}(t_i) = \hat{\mathbf{C}}^{gs}(t_{i}) - \hat{\mathbf{C}}^{gs}(t_{N})$. Moreover, the novel event-based residual loss avoids the influence of explicit event thresholds on reconstruction quality.

\begin{figure*}[t]
    \centering
    \includegraphics[width=\textwidth, trim=25 60pt 50 188pt, clip]{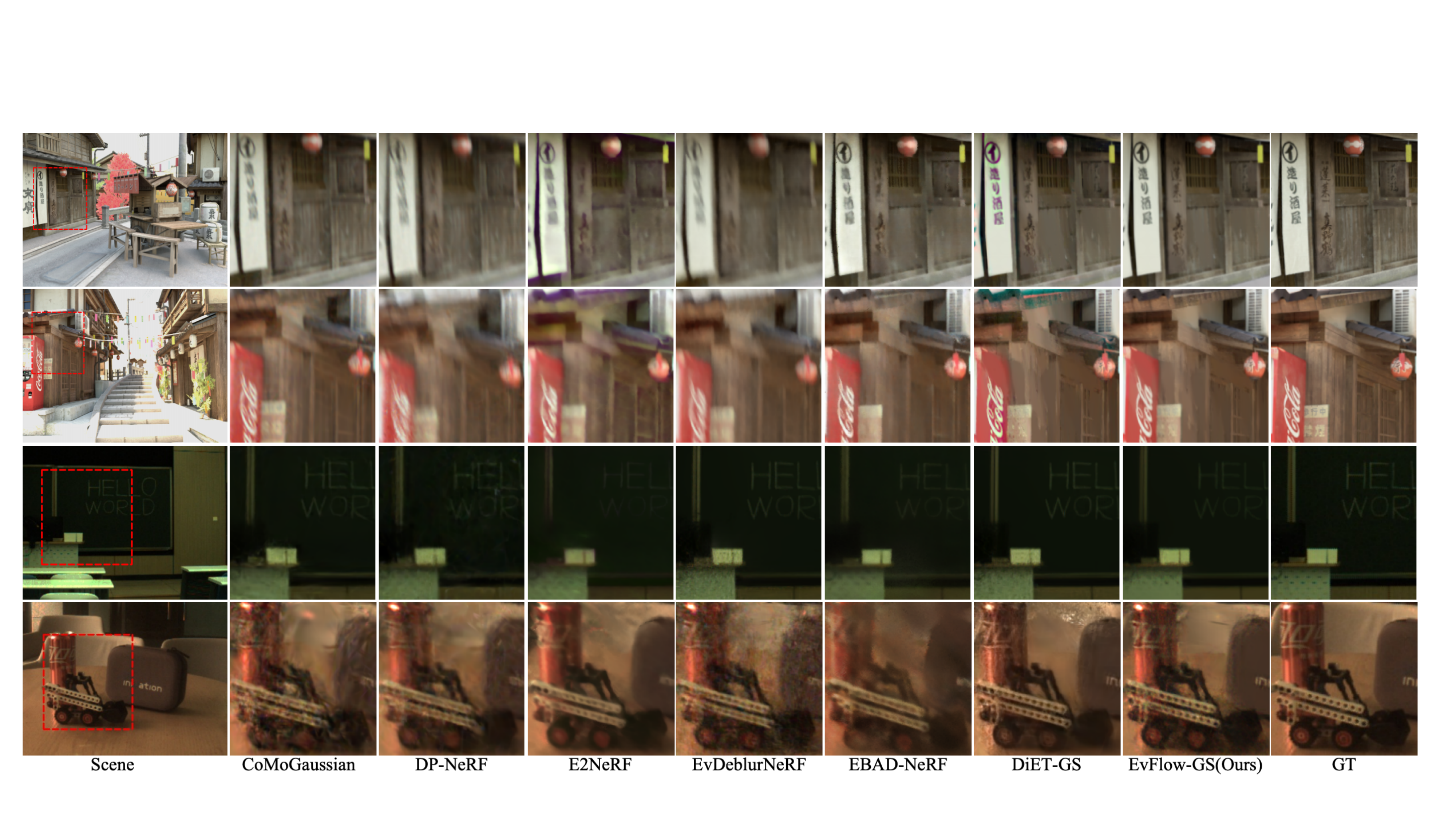}
    \caption{Qualitative comparison on datasets. The first and second rows show scenes from the synthetic dataset, while the third and fourth rows display results from two distinct real-world datasets. The results demonstrate the superiority of our approach in preserving details and reducing artifacts.}
    \label{fig:3}
\vspace{-\baselineskip}
\end{figure*}
\vspace{-2pt}
\begin{figure}[t]
\centering
\includegraphics[width=\linewidth, clip, trim=1 515pt 855 1pt]{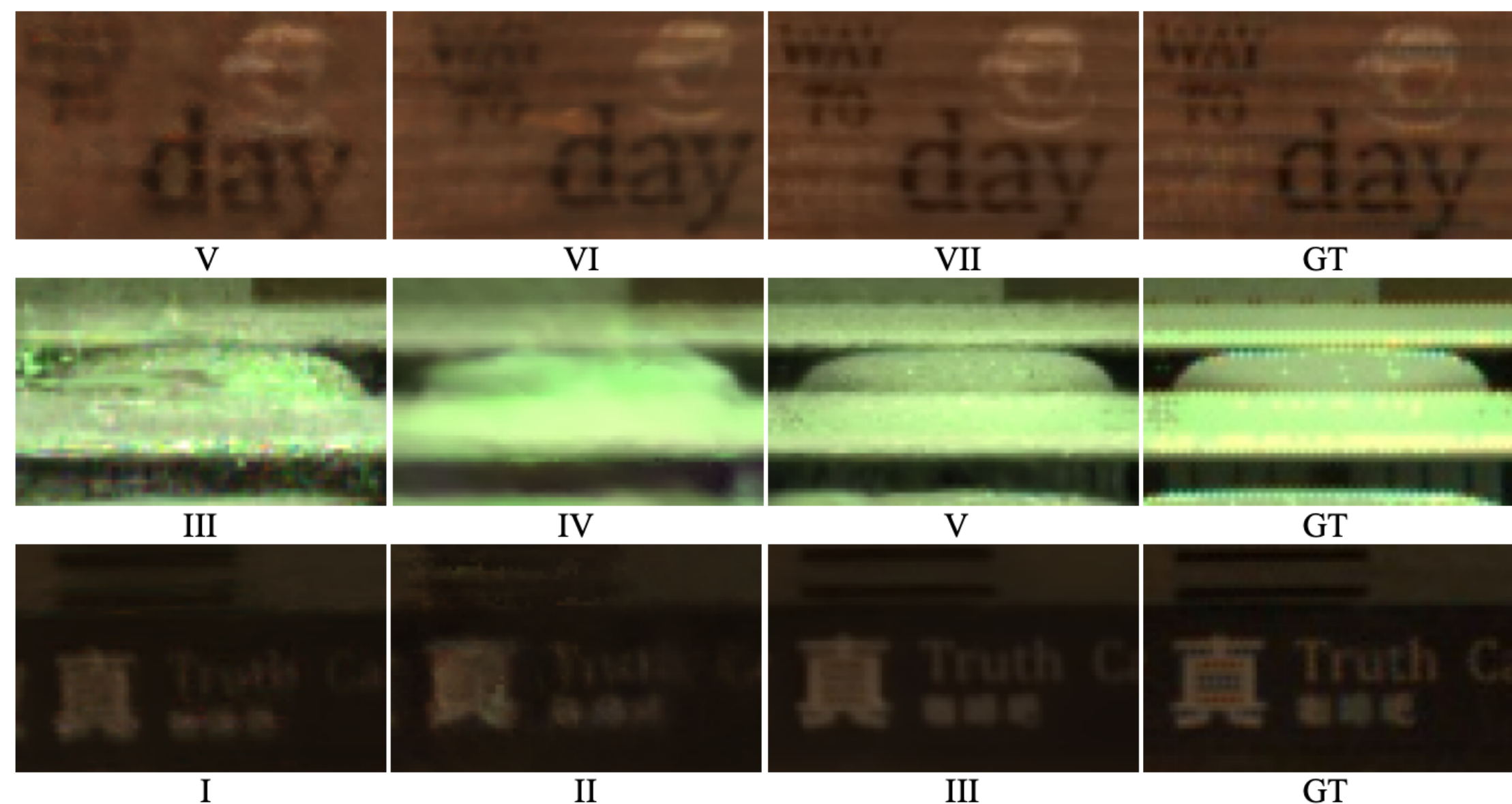}
\caption{Visual ablation experiment}
\label{fig:4}
\vspace{-\baselineskip}
\end{figure}
\vspace{-5pt}
\subsection{Learnable Double Integral Constraint.}
\label{sec:4.5}
Equation \eqref{eq:5} can recover sharp details but lacks supervision in areas where no events are triggered. Moreover, since the only color guidance comes from $\mathcal{L}_{\text{blur}}$, the color response of the event camera is poorly constrained, allowing 3DGS to add unnecessary texture details as long as the final blurred rendering looks correct. Inspired by~\cite{zhang2022unifying}, we jointly optimize a LDI network and 3DGS to mutually enhance the fidelity of deblurred texture details.\\
\textbf{LDI Network.}\hspace{1em}The LDI network is designed to be lightweight, consisting of five 2D convolutional layers. To address the issue of inaccurate double integration in EDI, we obtain event double integral through \( 
E(t_i, \mathcal{T}) 
\approx w_1 \, \text{LDI} \, (P(\mathcal{E}_{[t_i, t_0]})) + w_2 \, \text{LDI} \, (P(\mathcal{E}_{[t_i, t_{2N}]}))\)
where $\mathrm{LDI}(\cdot)$ is used for approximating $ E\!\left(0,\,[0,\, t_e - t_s]\right)$, $w_1 = (t_i - t_0)/T$, 
$w_2 = (t_{2N} - t_i)/T$ are weights, 
and $\mathcal{P}(\cdot)$ denotes the preprocessing event operator for 
time shift/flip and polarity reversal. We can reconstruct sharp frames $\{\hat{\mathbf{C}}_i^{ldi}\}_{i=0}^{2N}$ uniformly across \(\mathcal{T}\) using Eq. \eqref{eq:3}, and supervise LDI with both the blur loss and event-based loss: 
\begin{equation}
\mathcal{L}^{ldi} = \mathcal L_{blur}^{ldi} +\lambda\mathcal{L}_{ev}^{ldi}. 
\end{equation}
The event-based loss provides comprehensive temporal supervision for the LDI network, enhancing its generalization capability by enforcing consistency with the high-temporal-resolution event data. Moreover, convolution denoises event streams, which enhances the LDI's performance in edge recovery and noise removal. This enhancement, in turn, supplies 3DGS with color priors that contain more realistic and richer details for supervision.\\ 
\textbf{Joint Optimization.}\hspace{1em}In Sec.~\ref{sec:4.1} and \ref{sec:4.5} we can obtain the image sequence $\{\hat{\mathbf{C}}_i^{ldi}\}_{i=0}^{2N}$ from LDI network and $\{\hat{\mathbf{C}}_i^{gs}\}_{i=0}^{2N}$ rendered from the 3DGS. While both sequences represent the same scene, the joint optimization loss \(\mathcal{L}_{joint}\) can be formulated as 
\begin{equation}
\mathcal{L}_{\text{joint}} = \frac{1}{2N+1} \sum_{i=0}^{2N} \mathcal{L}\big( \hat{\mathbf{C}}_i^{ldi} , \hat{\mathbf{C}}_i^{gs} \big). 
\end{equation}
The LDI network compensates for the lack of motion features and reliable color in blurry inputs by providing direct supervision to 3DGS, while the multi-view consistency of 3DGS benefits coherent outputs from the LDI network. Through this mutual guidance, the framework effectively eliminates texture blurring, thereby significantly improving the clarity of reconstruction.

To sum up, the overall optimization objective for our EvFlow-GS framework enables the joint optimization of the LDI network, learnable poses, and 3DGS in a unified framework, and is formulated as:
\begin{equation}
\min \;
\mathcal{L}^{gs}_{blur}
+ \lambda \mathcal{L}^{gs}_{ev}
+ \lambda \mathcal{L}_{res}
+ \mathcal{L}^{ldi}
+ \mathcal{L}_{joint}.
\end{equation}

\section{Experiments}
\subsection{Experimental Setups}
\noindent\textbf{Implementation Details.}\hspace{1em}Our code is based on Bad-Gaussian~\cite{zhao2024badgaussians} on one NVIDIA A800 GPU, leveraging the Adam optimizer for joint optimization of LDI, pose estimation, and 3DGS. We train the network for 40,000 iterations. We choose $N = 3$ for all datasets, enabling each viewpoint to reconstruct a temporally uniform
sequence of 7 frames. The loss weights are set to $\lambda = 0.1$ and $\lambda_{SSIM} = 0.2$.\\
\noindent\textbf{Dataset.}\hspace{1em}
We evaluate our EvFlow-GS on the datasets proposed by EBAD-NeRF~\cite{qi2024deblurring} and on an additional real-world dataset from E3NeRF~\cite{qi2024e3nerfefficienteventenhancedneural}. EBAD-NeRF Dataset~\cite{qi2024deblurring} includes five synthetic scenes and two real-world scenes. Real-World-Challenge Dataset~\cite{qi2024e3nerfefficienteventenhancedneural} includes four real scenes with heavier blur and noisier events than EBAD-NeRF Dataset. For all datasets, except E2NeRF and DiET-GS, we use EDI-recovered sharp images at the exposure midpoint as inputs to COLMAP to re-estimate camera poses.
\\
\noindent\textbf{Baselines.}\hspace{1em} We compare EvFlow-GS with three categories of baselines: 
(i) RGB/Event deblurring + 3DGS, i.e., MPRNet~\cite{Zamir2021MPRNet}, EDI~\cite{zhang2022unifying}, EFNet~\cite{sun2022event}; 
(ii) RGB-only deblurring reconstruction, i.e., BAD-Gaussian~\cite{zhao2024badgaussians}, DP-NeRF~\cite{Lee_2023_CVPR}, Deblurring 3DGS~\cite{lee2024deblurring}, CoMoGaussian~\cite{lee2025comogaussian}; 
(iii) RGB+Event reconstruction, i.e., E2NeRF~\cite{qi2023e2nerf}, EvDeblurNeRF~\cite{Cannici_2024_CVPR}, EBAD-NeRF~\cite{qi2024deblurring}, DiET-GS~\cite{lee2025diet}.
For the Real-World-Challenge Dataset, we further include quantitative comparison with E3NeRF~\cite{qi2024e3nerfefficienteventenhancedneural}, whose code is not publicly available.
\begin{table}[t]
\centering
\caption{Ablation study on EvFlow-GS.}
\small
\vspace{-3.0mm}
\renewcommand{\arraystretch}{0.95} 
\setlength{\tabcolsep}{3pt}
\resizebox{\columnwidth}{!}{
\begin{tabular}{c|ccccccc|ccc}
\toprule
\multirow{2}{*}{ID}
& \multirow{2}{*}{$\mathcal{L}_{blur}^{gs}$}
& \multirow{2}{*}{$\mathbb{L}_{ev}$} 
& \multirow{2}{*}{$\mathbb{L}_{ev}^w$} 
& \multirow{2}{*}{$\mathcal{L}_{res}$} 
& \multicolumn{2}{c}{$\mathcal{L}_{ldi}$} 
& \multirow{2}{*}{$\mathcal{L}_{joint}$}
& \multicolumn{3}{c}{1st Real-World} \\
\cline{6-7} \cline{9-11}
& & & & 
& $\mathcal{L}_{blur}^{ldi}$ & $\mathcal{L}_{ev}^{ldi}$ 
& 
& PSNR$\uparrow$ & SSIM$\uparrow$ & LILPS$\downarrow$ \\
\midrule

I   & \checkmark &  &  &  &  &  &  
& 29.72 & 0.8475 & 0.1633 \\

II  & \checkmark & \checkmark &  &  &  &  &  
& 30.59 & 0.8851 & 0.1601 \\

III & \checkmark & \checkmark & \checkmark &  &  &  &  
& 31.37 & 0.8816 & 0.1492 \\

IV  & \checkmark &  &  & \checkmark &  &  &  
& 30.72 & 0.8752 & 0.1543 \\

V   & \checkmark & \checkmark & \checkmark & \checkmark &  &  &  
& \underline{31.85} & 0.9010 & \underline{0.1407} \\

VI  & \checkmark & \checkmark & \checkmark & \checkmark & \checkmark &  & \checkmark
& 31.80 & \underline{0.9049} & 0.1410 \\

VII & \checkmark & \checkmark & \checkmark & \checkmark & \checkmark & \checkmark & \checkmark
& \textbf{32.92} & \textbf{0.9110} & \textbf{0.1315} \\
\bottomrule
\end{tabular}
}
\label{table:3}
\vspace{-\baselineskip}
\end{table}
\vspace{-3pt}
\subsection{Experimental Results}
\noindent\textbf{Quantitative Comparisons.} We show the quantitative results in Table~\ref{tab:2}. Our EvFlow-GS outperforms baselines by significant margins in PSNR, SSIM, and LPIPS across datasets, demonstrating the effectiveness of our framework in exploiting event stream information.\\
\noindent\textbf{Qualitative Comparisons.} Qualitative comparisons on one synthetic dataset and two real-world datasets are shown in Fig.~\ref{fig:3}. Several observations can be made: (i) Previous works~\cite{Cannici_2024_CVPR,lee2025diet,qi2023e2nerf,qi2024deblurring} that rely solely on event accumulation to optimize 3DGS will generate lots of blurry artifacts at the edges. This underscores the necessity of our proposed warped event-based photometric constraint. For example, in the 4th row, it was observed that the third baseline resulted in grid artifacts around the cola and car, while EvFlow-GS restored more accurate textures. (ii) EvFlow-GS can obtain textured details close to GT in non-edge and low-light regions, demonstrating the effectiveness of the novel residual supervision. For example, in the third row of the podium area, our reconstruction shows significantly fewer black artifacts compared to baselines. (iii) Our optimization framework provides a stronger color supervision for 3DGS, thereby recovering more accurate textures and sharper details. As shown in 1st-2nd rows, due to improper pose representation, the triple-cascade baseline methods E2NeRF and DiET-GS exhibit color distortion in roof and text regions, while DiET-GS and EvDeblurNeRF using EDI supervision produces erroneous color jitter and black artifacts.
\vspace{-3pt}
\subsection{Ablation Study}
This section ablates each component of EvFlow-GS, with all evaluations conducted on the first real-world dataset. The results are shown in Table~\ref{table:3} and Fig.~\ref{fig:4}.\\
\noindent\textbf{Event-based Supervision.}\hspace{1em}While adding event accumulation supervision \(\mathcal{L}_{ev}\) in Eq.~(\ref{eq:5}) improves the PSNR by +0.87 dB, the rendered edges remain notably blurry and even fail to reconstruct letters, as shown in the 1st row of Fig.~\ref{fig:4}. This is because the uncorrected event accumulation forms a blurred distribution along incorrect edges, which misleads the 3DGS into learning wrong reconstructions. We warp events along correct trajectories to align them on sharp edges, providing clear and reliable supervisory signals to the model. The 1st row of Fig.~\ref{fig:4} demonstrates that adding \(\mathbb{L}_{ev}^w\) can achieve a further performance improvement of +0.78 dB in PSNR, and aid in fine-grained deblurring.\\
\noindent\textbf{Novel Residual  Supervision.}\hspace{1em}Although the warped event stream provides a sharper edge prior, it contains significant event noise in non-edge and low-light regions, preventing the rendering of correct textures and colors. The log-difference issue discussed in Sec.~\ref{1} further leads to training instability, ultimately resulting in failed reconstruction of low-intensity regions, visualized in the 2nd row of Fig.~\ref{fig:4}. In contrast, relying solely on \(\mathcal{L}_{res}\) mitigates the impact of event noise, thereby reducing artifacts and reconstructing clear colors in low-light regions. However, due to the data gap between pre-trained model and our application scenario, the rendered edges are not sharp. By integrating both \(\mathcal{L}_{ev}\) and \(\mathcal{L}_{res}\), our method achieves both faithful color reproduction and well-defined structural details, as evidenced in the 2nd row of Fig.~\ref{fig:4}.\\
\noindent\textbf{Joint Optimization.}\hspace{1em}The 3nd row of Fig.~\ref{fig:4} further demonstrates that Insufficient color supervision also introduces artifacts.
As shown in Table~\ref{table:3} and Fig.~\ref{fig:4}, \(\mathcal{L}_{joint}\) significantly enhances all 3DGS metrics, notably achieving a +1.07dB PSNR gain. Our framework enables a beneficial cycle where the LDI network enhances the 3DGS with sharp textures and colors, while the multi-view consistency from the 3DGS feeds back to refine the LDI network's outputs. Ultimately, this enables the NVS of 3DGS to render high quality details. However, constraining LDI with only \(\mathcal{L}_{blur}^{ldi}\) leads to a trivial mapping from ``\({\mathbf{C}}_{gt}+\mathcal{E}_{\mathcal{T}}\)" to \(\mathbf{C}_{gt}\). In contrast, \(\mathcal{L}_{ev}^{ldi}\) enhances the generalization and performance of LDI, showing a +1.12dB increase in PSNR. Our framework ultimately allows 3DGS to output results with precise colors and clear details as shown in the 1st row of Fig.~\ref{fig:4}, while the under-constrained LDI leads to over-smoothed details in the 3DGS rendering.
\section{Conclusion}
We present EvFlow-GS, which effectively fuses event streams with optical flow priors.
By jointly optimizing the  an end-
to-end learnable double integral network, learnable camera poses, and 3D Gaussian Splatting, our approach suppresses artifacts and recovers sharper texture details, resulting in high-quality reconstruction.
\vspace{+2pt}
\section*{Acknowledgments}
\label{sec:Acknowledgments}
{\setlength{\emergencystretch}{5em}
\tolerance=2000       
\sloppy               
This work was supported by the Sichuan Science and Technology Program (No. 2025ZDZX0093), Open Research Fund of the State Key Laboratory of Brain-Machine Intelligence, Zhejiang University (No. BMI2500014), Sichuan Province Innovative Talent Funding Project for Postdoctoral Fellows, National Major Scientific Instruments and Equipments Development Project of National Natural Science Foundation of China (No. 62427820) and the Science Fund for Creative Research Groups of Sichuan Province Natural Science Foundation (No. 2024NSFTD0035).

\bibliographystyle{IEEEbib}
\bibliography{main}

\end{document}